\documentclass[11pt]{article}

\usepackage[final]{acl}

 \usepackage{amsmath} 
\usepackage{times}
\usepackage{latexsym}
\usepackage{enumitem}
\usepackage[T1]{fontenc}
\usepackage{xcolor}
\usepackage{colortbl}
\usepackage{multirow} 
\usepackage[utf8]{inputenc}

\usepackage{microtype}

\usepackage{inconsolata}
\usepackage{amssymb}
\usepackage{booktabs}
\usepackage{graphicx}
\usepackage{subfigure}

\title{Perplexity-Aware Data Scaling Law: Perplexity Landscapes Predict Performance for Continual Pre-training}
\author{
 \textbf{Lei Liu\textsuperscript{1,2}},
 \textbf{Hao Zhu\textsuperscript{1}},
 \textbf{Yue Shen\textsuperscript{1}},
  \textbf{Zhixuan Chu\textsuperscript{2,$\dagger$}},
 \textbf{Jian Wang\textsuperscript{1}},
 \textbf{Jinjie Gu\textsuperscript{1}},
 \textbf{Kui Ren\textsuperscript{2}}
\\
\\
 \textsuperscript{1}Ant Group,
 \textsuperscript{2}Zhejiang University
\\
\\
\textbf{Email:} \href{liulei1497@gmail.com}{liulei1497@gmail.com}, \textbf{Correspondence:} \href{zhixuanchu@zju.edu.cn}{zhixuanchu@zju.edu.cn} 
}

\begin{document}
\maketitle
\begin{abstract}
Continual Pre-training (CPT) serves as a fundamental approach for adapting foundation models to domain-specific applications. Scaling laws for pre-training define a power-law relationship between dataset size and the test loss of an LLM. However, the marginal gains from simply increasing data for CPT diminish rapidly, yielding suboptimal data utilization and inefficient training. To address this challenge, we propose a novel perplexity-aware data scaling law to establish a predictive relationship between the perplexity landscape of domain-specific data and the test loss. Our approach leverages the perplexity derived from the pre-trained model on domain data as a proxy for estimating the knowledge gap, effectively quantifying the informational perplexity landscape of candidate training samples. By fitting this scaling law across diverse perplexity regimes, we enable adaptive selection of high-utility data subsets, prioritizing content that maximizes knowledge absorption while minimizing redundancy and noise. Extensive experiments  demonstrate that our method consistently identifies near-optimal training subsets and achieves superior performance on both medical and general-domain benchmarks.
\end{abstract}

\section{Introduction}
Large language models (LLMs) have demonstrated impressive capabilities across a wide range of domains \citep{liu2024survey}. However, their general-purpose pre-training objectives often leave them ill-suited for specialized applications such as healthcare, where domain-specific knowledge, precise terminology, and structured reasoning are critical. To bridge this gap, Continual Pre-Training (CPT) has emerged as a dominant paradigm \citep{que2024d}: by further pre-training a general-purpose LLM on domain-specific corpora, models can internalize nuanced medical concepts, factual knowledge, and domain-typical reasoning patterns, thereby improving performance on downstream tasks.

Despite its success, CPT remains largely guided by heuristic data practices, with limited understanding of how data characteristics influence learning dynamics \citep{wang2025learning,chen2025towards}. A key challenge is the inefficiency of scaling data under continual pre-training. Classical scaling laws exhibit power-law predictive relationship between loss and dataset size, but they assume that each token contributes equally to learning \citep{hoffmann2022training}. In practice, domain-specific corpora exhibit high levels of redundancy and noise, and vary significantly in conceptual density. For instance, biomedical literature may contain long passages that restate known facts, while clinical notes often include unstructured or repetitive entries. As a result, simply increasing the amount of training data leads to sharply diminishing returns. This observation underscores the need for data-informed strategies that move beyond raw data volume and instead emphasize the quality, diversity, and informational value of training samples. 

This breakdown calls for a shift from purely quantity-driven paradigms to data-centric strategies that explicitly account for  sample effectiveness in CPT \citep{yu2024mates,engstrom2024dsdm}. Rather than treating all domain texts equally, we argue that one should prioritize instances that most effectively reduce the model’s uncertainty, particularly those that target its most salient knowledge gaps. A natural question then arises: how can we quantify such gaps in a manner that is both computationally efficient and strongly correlated with downstream performance improvements?

In this work, we propose that the answer lies in the model’s own uncertainty signal: \emph{perplexity} \citep{ankner2024perplexed}. We introduce the concept of \emph{perplexity landscapes}, fine-grained distributions of model perplexity over streaming domain data, as a powerful diagnostic tool for characterizing the knowledge frontier between general and domain-specific expertise. Crucially, we observe that the shape of these landscapes at early stages of CPT strongly correlates with eventual fine-tuning performance, suggesting that initial perplexity encodes actionable information about data utility.

Building on this insight, we derive a novel \emph{perplexity-aware data scaling law} that establishes a predictive functional relationship between statistics of the initial perplexity distribution (e.g., mean, variance, tail mass) and final task performance after CPT. Unlike traditional scaling laws based solely on data volume, our law incorporates intrinsic model responses to individual data points, enabling adaptive selection of training subsets that maximize knowledge absorption while filtering out redundant, overly difficult, or noisy samples.

Our method requires only a single forward pass over unlabeled domain data using the frozen initial model, making it efficient and scalable. By fitting the scaling law on small pilot batches, we can estimate the expected return of larger data subsets and select those predicted to yield optimal performance. This enables principled, model-informed data curation without requiring labeled examples or expensive retraining loops. We validate our approach across medical and general benchmarks. Results show that perplexity landscapes consistently identify near-optimal data subsets, achieving a superior improvement with perplexity as a proxy for knowledge acquisition. Our contributions are threefold:
\begin{itemize}[leftmargin=*]
    \item We introduce perplexity landscapes as a diagnostic tool for CPT, where perplexity distributions are strongly predictive of downstream performance, providing a window into the evolving knowledge frontier during specialization.
    \item We propose a novel perplexity-aware data scaling law to link statistics of data perplexity to final CPT performance, which moves beyond only scaling data volume.
    \item We develop an efficient model-aware data selection framework to identify high-value training samples. Our method enables scalable data curation for achieving superior performance with less data and demonstrates consistent gains across medical and general benchmarks.
\end{itemize}

\section{Related Work}

\paragraph{Scaling Law} A growing body of research has established that the performance of large language models (LLMs) follows predictable scaling laws with respect to key resources such as model size, training compute, and dataset size \citep{kaplan2020scaling, hoffmann2022training}. In particular, numerous studies have demonstrated that model performance improves according to a power-law relationship as the number of parameters or the volume of training data increases \citep{kaplan2020scaling, hoffmann2022training}, enabling principled extrapolation from small-scale experiments to large-scale deployments. These scaling laws provide a theoretical foundation for optimizing training efficiency and guide decisions in model design and data budget allocation. More recently, scaling laws have been extended beyond parameter and data size to encompass more nuanced factors, such as data mixture proportions. For instance, \citet{que2024d} proposed a data-mixing scaling law that predicts performance based on the composition of multi-domain training sets, offering guidance for curriculum and domain-adaptive pre-training.

\section{Data-centric Scaling Law by Perplexity}
Unlike pretraining, the primary challenge in CPT lies in balancing knowledge retention (the preservation of the model’s previously acquired capabilities) with knowledge acquisition (effective adaptation to and integration of new information). Consequently, data selection becomes even more critical in CPT. 

\subsection{Motivation}
\paragraph{Motivation-1: Marginal Gain of Scaling Data Diminishes During CPT.}
Classical scaling laws \citep{kaplan2020scaling,hoffmann2022training} works with a potential assumption, \textit{uniformly informative data}. For example, empirical scaling law \citep{hoffmann2022training} fits a parametric loss function, building the systematic, predictable connections among model size, training dataset size, and the model’s final performance. 
\begin{equation}
\hat{L}(N, D) \triangleq E+\frac{A}{N^\alpha}+\frac{B}{D^\beta}
\end{equation}
where $N$ denotes the parameters and $D$ represents the number of training tokens. $\alpha$ is parameter scaling exponent and $\beta$ is dataset scaling exponent. $E$ corresponds to the entropy of natural text. These laws quantify how performance improves as the attributes are "scaled up", providing a foundational framework for guiding LLM development. 

However, during CPT, the marginal gain from simply increasing dataset size diminishes, where data quantity becomes inadequate predictor for model performance due to the uncertainty from the base model and significantly varying data distribution. This suggests that conventional scaling laws, which focus solely on quantity, may be insufficient to capture the dynamics of model refinement in later training stages. Noticing that factors related to data effectiveness (\textit{e.g.}, knowledge density, topical relevance, and factual accuracy) emerge as primary drivers of further performance improvement, there is a compelling need to extend the scaling law framework to incorporate a dataset importance weighting dimension, potentially yielding a two-axis formulation that jointly models the effects of dataset size and dataset importance on loss reduction. This motivates the question:

\emph{Given a pre-trained model, under the condition of a fixed number of training texts, how to determine the optimal training data subset for continual pre-training?}

\paragraph{Motivation-2: Perplexity Landscapes Predict CPT Performance.} To select the effective data, perplexity provides a natural metric~\citep{ankner2024perplexed}: sequences with low PPL are redundant, while those with high PPL are likely noisy or incomprehensible, both yield diminishing learning returns. The most effective data lie in a "sweet spot" of moderate perplexity, which formalizes the intuition that the most valuable data is neither too similar nor too diverse to the model.

Starting with trained models under different data subsets of varying PPL distributions, experimental results are used to fit an empirical estimator for determining the optimal perplexity range. This formulation extends scaling laws by incorporating perplexity statistics, \textit{i.e.}, the mean and variance. Given the fixed tokenizer and corpus, such statistics become dimensionless across model scales. 

% thereby leading to a pure power-law scaling relationship
% Such an approach could provide a more accurate predictive tool for designing data curation and filtering strategies in the high-performance regime of LLM training.

\subsection{Perplexity-Aware Data Scaling Law}  
\subsubsection{Scaling Law Formula}
Focusing on the data-centric term in \citep{kaplan2020scaling,hoffmann2022training}, we utilize the following functional form over the both dataset size and importance:
\begin{equation}
\hat{L}(Q, D) \triangleq E+\frac{D_c}{Q* D^{\alpha_{D}}},
\end{equation}
where $D_c$ and $\alpha_{D}$ are hyper-parameters to be fitted. $E$ corresponds to the entropy of natural text. Here, we omit the starting status of the pre-training model because the following perplexity descriptor contains such information. The above formulation typically assumes homogeneous data distributions, treating $Q$ as a constant or implicit factor. Here, we consider parameterize the dataset importance term using basic informativeness measurement. 

\begin{figure*}[!t]
    \centering
    \subfigure[Qwen3-0.6B-Base]{
        \includegraphics[width=0.48\textwidth]{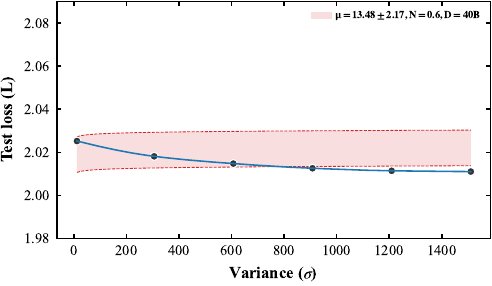}
        \label{fig:sub2}
    }
    \subfigure[Qwen3-14B-Base]{
        \includegraphics[width=0.48\textwidth]{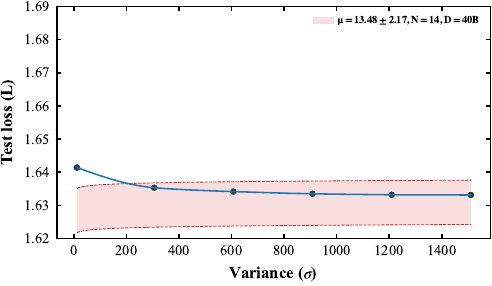}
        \label{fig:sub3}
    }
    \caption{\centering  Interdependence between loss and $\mu$ ($\sigma$).} % 整个图的总标题
    \label{fig:scale_vis_comp_without_interaction} % 整个图的标签，用于交叉引用
\end{figure*}

\begin{figure*}[!t]
    \subfigure[\centering Test Loss \textit{VS.} \textcolor{red}{PPL Distribution}]{
\includegraphics[width=0.48\textwidth]{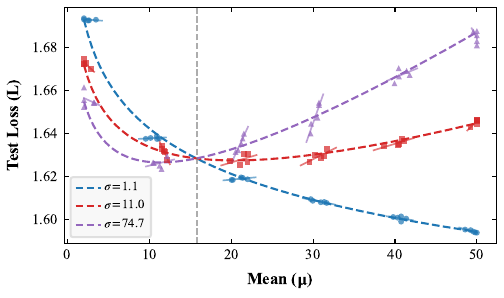}
        \label{fig:scaling-sub1}
    }
    \hfill % 子图3和子图4之间的水平均匀填充
    \subfigure[\centering Test Loss (Log) \textit{VS.} \textcolor{red}{FLOPs}]{
\includegraphics[width=0.48\textwidth]{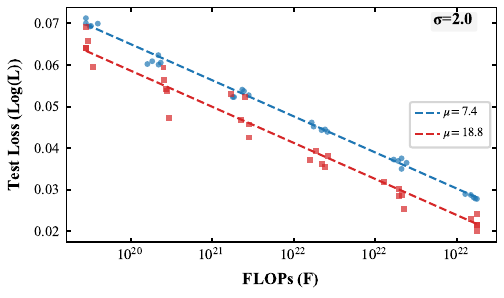}
\label{fig:scaling_law_formula_flops}
    }
\caption{\centering Fitted curves for scaling law based on Qwen3-14B-Base.} 
    \label{fig:scaling_law_formula}
\end{figure*}

Perplexity (PPL), as a token-level likelihood measure under a reference model, provides a fine-grained proxy for sample informativeness \citep{brown2020language,touvron2023llama}. We argue that summarizing the PPL distribution across a dataset via its statistic distribution (mean and variance) offers a principled and scalable importance indicator. In detail, the mean captures the average sample difficulty, while the variance reflects diversity, both of which can influence learning dynamics and generalization~\citep{zhang2025frame}. Incorporating these statistics allows for a more nuanced understanding of how data modulates model performance, enabling better predictions and resource allocation in large-scale training regimes.

Accordingly, let $\mu$ and $\sigma$ indicate the mean and variance of PPL distribution respectively. We instantiate the dataset importance term as $Q(\mu, \sigma) = \mu^{\alpha_\mu}*\sigma^{\alpha_\sigma}$, which models the joint influence of mean and variance in a scale-invariant manner. Here, $\alpha_\mu$ and $\alpha_\sigma$ are hyper-parameters. By enriching the scaling law with interpretable, data-driven signals, we have:
\begin{equation}
\hat{L}(\mu, \sigma, D) \triangleq  E + \frac{D_c}{(\mu^{\alpha_\mu}*\sigma^{\alpha_\sigma})*D^{\alpha_{D}}}.
\label{eq:ppl_scaling_law_basic}
\end{equation}

Empirical experiments (Figure~\ref{fig:scale_vis_comp_without_interaction}) indicate that the relation between loss and $\mu$ ($\sigma$) is not strictly monotonic, while these two variables exhibit a measurable interdependence. Thus, we introduce the minimal multiplicative interaction to extend Eq.~\ref{eq:ppl_scaling_law_basic} as follows:
\begin{equation}
    \alpha_\mu(\sigma) = \alpha_0 + \alpha_1\sigma, \quad \alpha_\sigma(\mu) = \beta_0 + \beta_1\mu,
\label{eq:ppl_scaling_law_new}
\end{equation}
where $\alpha_1$ and $\beta_1$ are relationship variables between $\mu$ and $\sigma$. Such transformation results in the following format of perplexity-aware data scaling law:
\begin{equation}
\hat{L}(\mu, \sigma, D) \triangleq  E + \frac{D_c}{(\mu^{\alpha_\mu(\sigma)}*\sigma^{\alpha_\sigma(\mu)})*D^{\alpha_D}},
\label{eq:ppl_scaling_law_inter}
\end{equation}
which preserves the interpretable power-law structure. Besides, it incorporates the relationship decomposition between mean and variance of perplexity distribution, \textit{i.e.}, interdependence and independence decomposition:
\begin{equation}
L(\mu, \sigma, D) \triangleq  E + \frac{D_c}{\underbrace{\mu^{\alpha_0}\sigma^{\beta_0}}_{\text{Independence}}*\underbrace{\mu^{\alpha_1\sigma}\sigma^{\beta_1\mu}}_{\text{Interdependence}}*D^{\alpha_D}}.
\end{equation}
This format is strongly favored because it (i) conserves the power-law structure, (ii) keeps interactions simple, and (iii) collapses to the basic model when variables are independent.

\begin{figure*}[!t]
    \subfigure[Test Loss \textit{VS.} \textcolor{red}{Parameters} (Low PPL Range)]{
        \includegraphics[width=0.48\textwidth]{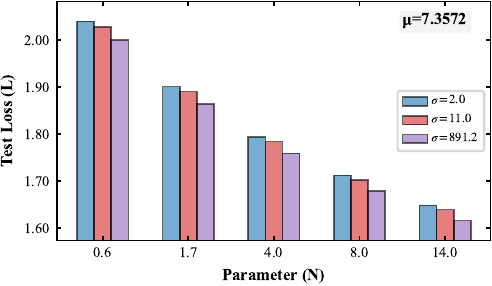}
        \label{fig:scaling-sub3}
    }
    \hfill % 子图3和子图4之间的水平均匀填充
    % 子图4：宽度与其他子图完全一致
    \subfigure[Test Loss \textit{VS.} \textcolor{red}{Parameters} (High PPL Range)]{
        \includegraphics[width=0.48\textwidth]{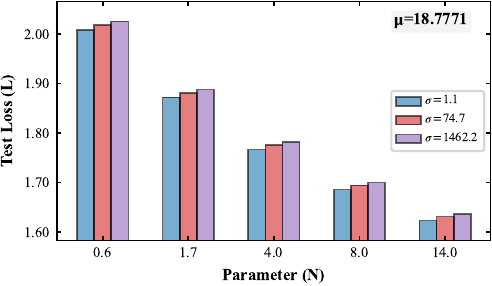}
        \label{fig:scaling-sub4}
    }
 \caption{\centering Performance changes with varying PPL distributions and Parameters.} 
    \label{fig:scaling_law_performance}
\end{figure*}

\begin{figure*}[t]
    \centering
    % $L=2.11\mu^{-0.019}\sigma^{-0.007+0.001\mu}N^{-0.065}D^{-0.001}$ 
    % 第二张子图
    \subfigure[Qwen3-0.6B-Base]{
        \includegraphics[width=0.48\textwidth]{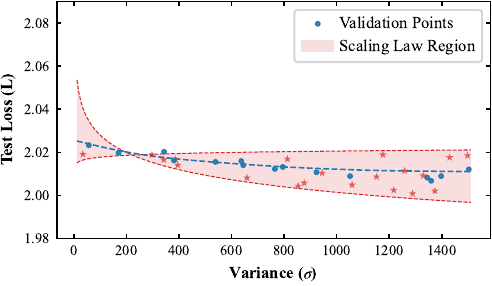}
        \label{fig:scale_vis_comp_sub1}
    }
    % 第三张子图
    \subfigure[Qwen3-14B-Base]{
        \includegraphics[width=0.48\textwidth]{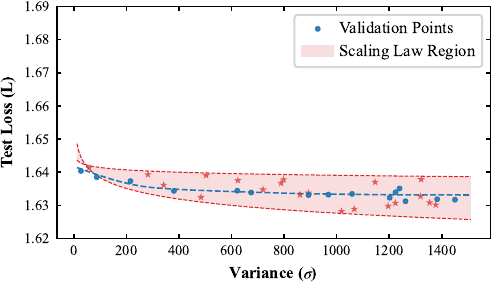}
        \label{fig:scale_vis_comp_sub2}
    }
    \caption{\centering Validation curves of scaling law.} % 整个图的总标题
    \label{fig:scale_vis_comp} % 整个图的标签，用于交叉引用
\end{figure*}

\subsection{Scaling Law Fitting}
\paragraph{Setting} We select medical domain as the target. Given the dataset from PubMed corpus, we firstly perform multiple bootstrap sampling to generate different subsets with varying $\mu$ and $\sigma$. Among these training subsets, $90\%$ subsets are used for fitting the scaling curve and $10\%$ subsets are for validation. The final fitting curves with sampled fitting points are shown in Figure \ref{fig:scaling_law_formula}. The validation results are shown in Figure~\ref{fig:scale_vis_comp}. The red region is obtained according to the fitted scaling law with $\mu=13.48\pm2.17$ and $\sigma$ varying from 25 to 1600 with $\mu=13.48\pm2.17$, which is the real PPL range of the dataset. It is observed that the fitted scaling law closely matches the validation points (blue) across different variances and maintains a consistent trend. This strong agreement confirms that the formula accurately captures the relationship between dataset importance, and performance.

\paragraph{Why Consider Interaction Term in Scaling Law?} Eq. \ref{eq:ppl_scaling_law_basic} assumes a monotonic relationship between the loss and each variable, while experimental results indicate that the relationship between the loss and the mean $\mu$ or standard deviation $\sigma$ is not purely monotonic. As shown in Figure~\ref{fig:scale_vis_comp_without_interaction}, fitting Eq. \ref{eq:ppl_scaling_law_basic}, the real samples (black points) are not strictly fall in the region we fitted (red area). Furthermore, since both $\mu$ and $\sigma$ are derived from the same PPL distribution, an intrinsic correlation between them may exist. Therefore, it is necessary to introduce interaction terms into the model to better capture their interdependent effects.

\paragraph{Why Perplexity Distributions Follow Power-Law Forms in Scaling Laws?}
Power-law behavior arises from language data’s inherent hierarchy. Natural language data exhibits scale-invariant hierarchical structure: linguistic units (tokens, phrases, documents, domains) follow a power-law distribution in their frequency and complexity. For example, Zipf’s law~\citep{zipf2013psycho} describes how word frequencies scale as \(f \propto r^{-s}\) (where $r$ is rank and \(s \approx 1\)), and document complexity (measured by syntactic depth or semantic ambiguity) similarly follows a power-law, where a small fraction of "high-complexity" documents drive most variation in PPL~\citep{wold2024estimating}. This hierarchy directly shapes \(\mu\) and \(\sigma\):

\begin{itemize}[leftmargin=*]
    \item Mean \(\mu\): As incremental data volume (\(D_{\text{new}}\)) scales, \(\mu\) converges to a domain-specific limit (\(\mu_0\)) because larger datasets increasingly sample the full range of linguistic complexity. The convergence rate follows a power law (\(\mu - \mu_0 \propto D_{\text{new}}^{-\alpha}\), \(\alpha > 0\)) because the remaining "unseen" complexity (driving deviations from \(\mu_0\)) is dominated by low-frequency, high-complexity data, whose contribution decays as a power of \(D_{\text{new}}\) (consistent with ~\cite{cagnetta2025learning}).
    \item Variance \(\sigma\): Variance quantifies diversity in data quality/complexity, which is inherently tied to the number of distinct subdomains (C) in \(D_{\text{new}}\). Each subdomain contributes a unique PPL sub-distribution. The total variance scales as \(\sigma \propto C^\gamma\) (\(\gamma > 0\)), \textit{i.e.}, a power law because subdomain complexity itself follows a power-law hierarchy.
\end{itemize}

\begin{figure*}[!t]
    \centering
    % 第二张子图
    \subfigure[PPL Descent Paths]{
        \includegraphics[width=0.48\textwidth]{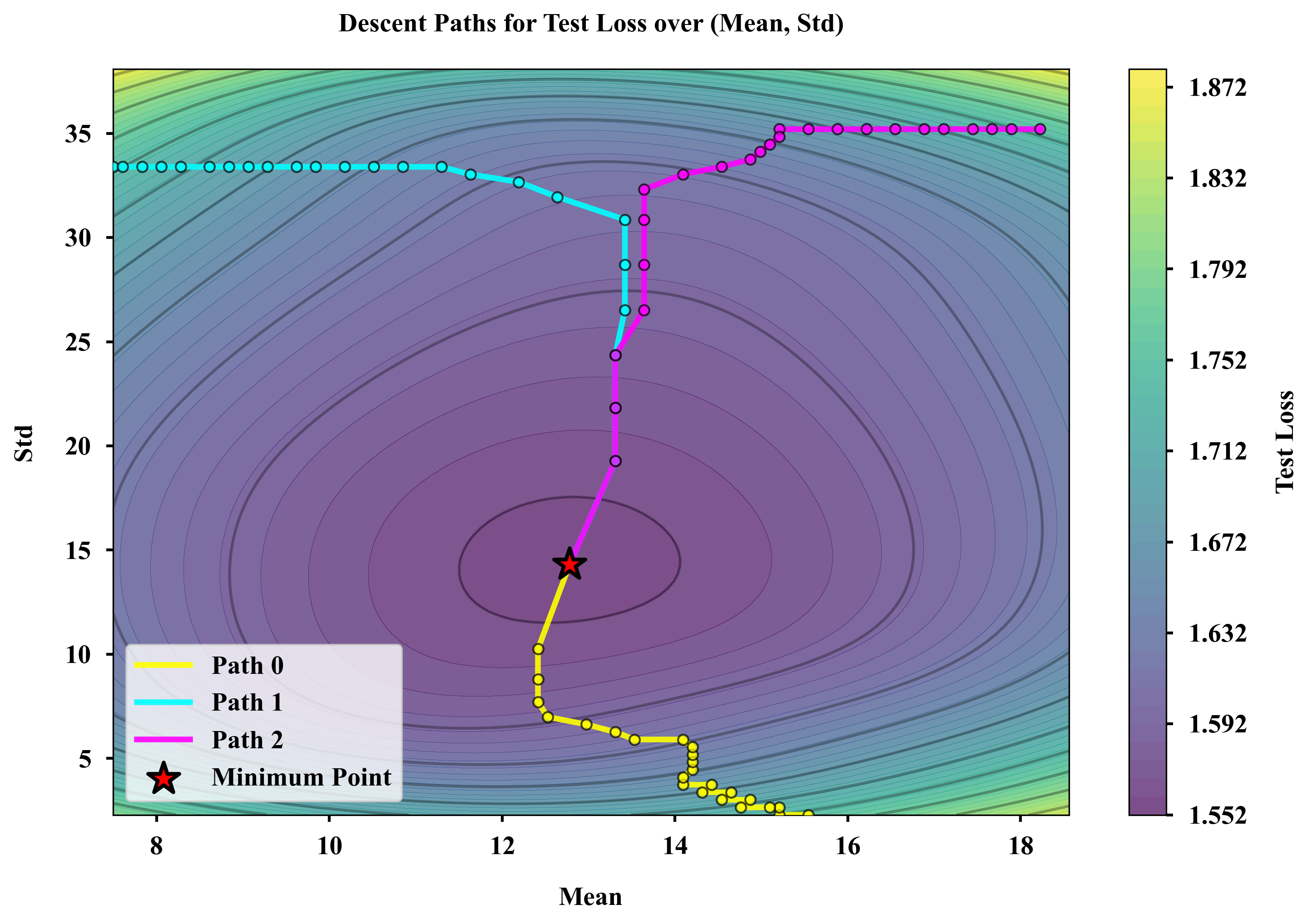}
        \label{fig:scale_vis_14B_sub2}
    }
    % 第三张子图
    \subfigure[Loss-PPL 3D Distribution]{
        \includegraphics[width=0.48\textwidth]{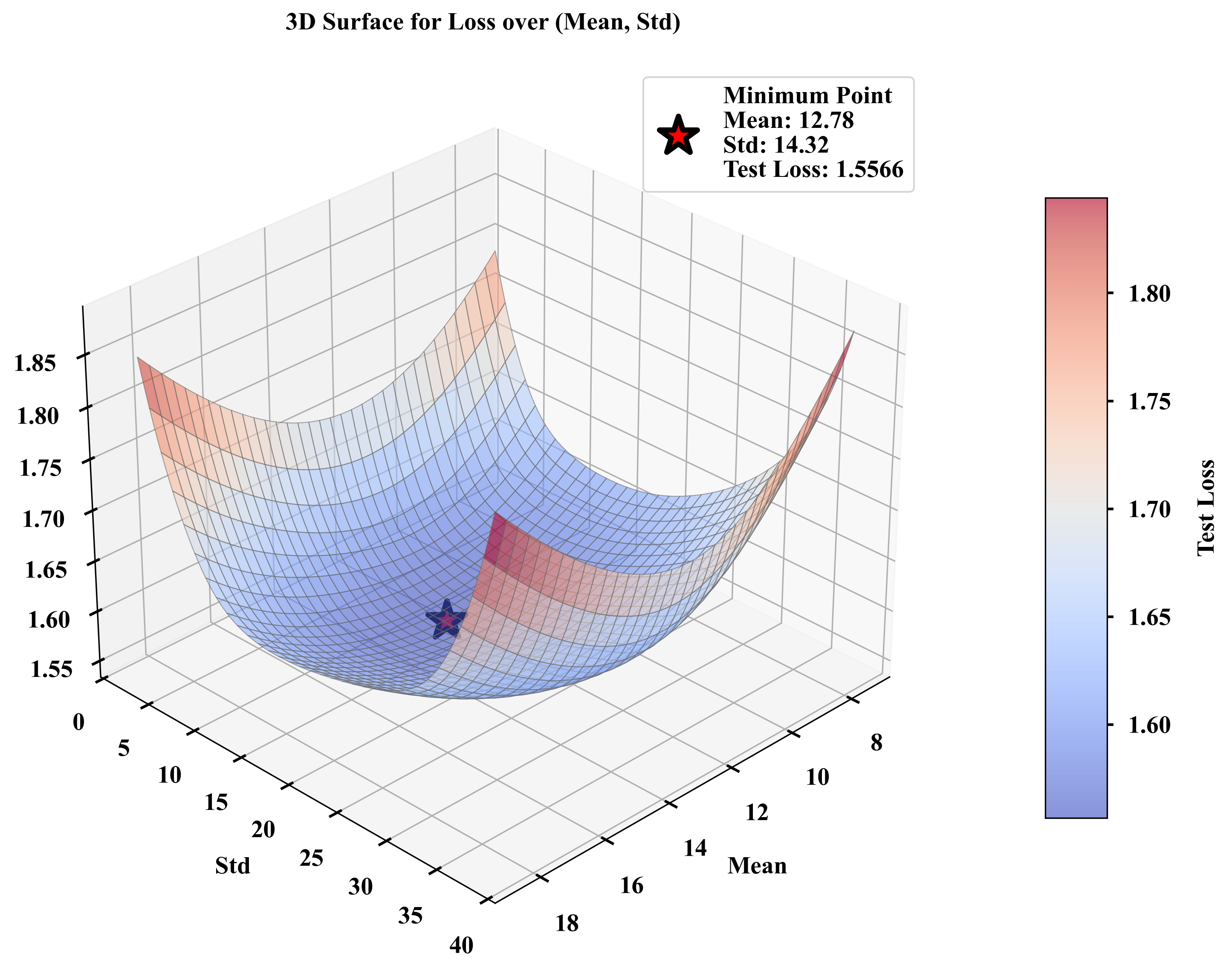}
        \label{fig:scale_vis_14B_sub3}
    }
    \caption{\centering Visualizations for perplexity landscape on Qwen3-14B-base model.} % 整个图的总标题
    \label{fig:scale_vis_14B} % 整个图的标签，用于交叉引用
\end{figure*}

\paragraph{How Perplexity Landscape Affects Performance?} From the data-centric perspective, the test loss of a LLM model can be predicted under the condition of data information, \textit{i.e.}, PPL mean ($\mu$), variance ($\sigma$), and dataset size ($D$). There are some key observations from Figure \ref{fig:scaling_law_formula} and \ref{fig:scaling_law_performance}.

\begin{itemize}[leftmargin=*]
    \item In Figure~\ref{fig:scaling-sub1}, perplexity distribution has a non-monotonic effect. There exists an optimal point for $\mu$ and $\sigma$. Higher $\sigma$ initially decreases loss by introducing useful diversity or hardness. However, beyond an optimal point $\mu$ would harm performance. Low-$\mu$ data benefits more from higher $\sigma$, while high-$\mu$ data suffers at very high $\sigma$.
    
    \item In Figure~\ref{fig:scaling_law_formula_flops}, with a moderate variance, lower mean yields worse convergence performance.
    
    \item In Figure~\ref{fig:scaling_law_performance}, test loss decreases with model size ($N$), but the magnitude depends on data characteristics. In a low PPL range (Figure~\ref{fig:scaling-sub3}), higher $\sigma$ results in lower test loss than lower $\sigma$ for the same number of parameters. In a high PPL range (in Figure~\ref{fig:scaling-sub4}), lower $\sigma$ enables significantly better scaling, whereas higher $\sigma$ limits.
\end{itemize}

\paragraph{Could Perplexity-Aware Scaling Law Generalize to New Points?}
To rigorously assess the generalization of the fitted scaling law, we evaluate its predictions on an independent validation set that is entirely disjoint from the data used for fitting. The validation configurations are sampled from the same underlying data distribution while exhibiting different perplexity (PPL) profiles, ensuring that the evaluation tests extrapolation to unseen points rather than simple interpolation. 

As shown in Figure~\ref{fig:scale_vis_comp}, the validation points (blue dots) closely follow the predicted trend and remain largely within the shaded scaling-law region for both Qwen3-0.6B-Base and Qwen3-14B-Base. Across the full range of variance, the empirical test losses align well with the power-law curve, and deviations are small and symmetric, indicating no systematic bias. This consistency between the model’s predictions and the independently obtained validation measurements demonstrates that the derived power-law relationship is both robust and reliable, and that it captures the dominant dependence of test loss on variance for these models.

\paragraph{PPL Landscapes Visualization}
Figure~\ref{fig:scale_vis_14B} visualizes how the test loss varies as a function of the mean and standard deviation of the PPL distribution, and how this relates to our scaling law.

In the contour plot (Figure \ref{fig:scale_vis_14B}(a)), the level sets of test loss form smooth, roughly elliptical basins in the (mean, std) plane. The three descent paths start from different initial PPL configurations but all move along the gradient of the loss surface toward the same low‑loss region, highlighted by the red star. This indicates that the loss is not determined by mean or variance alone; instead, it depends on their joint configuration, and different PPL profiles can converge to a common optimum predicted by the scaling law.

The 3D surface (Figure Figure \ref{fig:scale_vis_14B}(b)) makes this relationship explicit: test loss forms a bowl‑shaped surface over (mean, std), with a single, well‑defined minimum. Near this minimum, loss changes smoothly and approximately follows our power‑law scaling with respect to the PPL variance (for a given mean). Moving away from the optimum in either direction, by increasing or decreasing the mean or the variance—monotonically increases the loss, consistent with the scaling-law behavior observed in our 1D slices.

Thus, the PPL landscape analysis provides a geometric interpretation of the scaling law: the power-law relationship describes how test loss scales along the main descent directions in the (mean, std) space of perplexity.

\subsection{Distance-to-Optimum Selection}
Given a fitted perplexity-aware scaling law, we assume it identifies an optimal region of the data-perplexity distribution characterized by a target mean $\hat{\mu}$ and variance $\hat{\sigma}$. Our goal is to construct a continual pre-training subset whose empirical perplexity statistics approximate $(\hat{\mu}, \hat{\sigma}^{2})$ under a fixed token budget.

\paragraph{Distance-to-Optimum Objective} For a subset \(\mathcal{S}\), we measure the deviation from the optimal distribution by:
\begin{equation}
    J(\mathcal{S})
    = w_\mu \big(\mu(\mathcal{S}) - \hat{\sigma}\big)^2
    + w_\sigma \big(\sigma^2(\mathcal{S}) - \hat{\sigma}^{2}\big)^2,
\end{equation}
where \(w_\mu, w_\sigma > 0\) are weighting coefficients
(typically \(w_\mu = w_\sigma = 1\)). The Distance-to-Optimum Selection (DOS) problem can then be formulated as:
\begin{equation}
    \min_{\mathcal{S} \subseteq \mathcal{D}} \quad 
        J(\mathcal{S}) \text{s.t.} \quad 
        \sum_{c_j \in \mathcal{S}} |c_j|
        \leq T_{\text{budget}}.
\label{eq:selection-objective}
\end{equation}
Solving this problem exactly is intractable, therefore we adopt a greedy approximation.

\paragraph{Greedy Selection Algorithm} Initially, the whole corpus is chunked into $N$ random subsets \(\mathcal{D}=\{c_j\}_{j=1}^N\). Define the optimal subset as \(\mathcal{S} \leftarrow \emptyset \)  and data budge as \(\quad T \leftarrow 0\). To initialize a non-empty subset, choose the chunk whose perplexity is closest to $\hat{\mu}$, \textit{i.e.}, $j^*= \arg\min_{j} |p_j-\hat{\mu}|$ and add it to \(\mathcal{S}\).  The greedy expansion is conducted as following steps while \(T < T_{\text{budget}}\):
\begin{itemize}[leftmargin=*]
      \item For each candidate \(c_j \notin \mathcal{S}\) with \(T + |c_j|\le T_{\text{budget}}\),
      form \(\mathcal{S}' = \mathcal{S} \cup \{c_j\}\), update
      \(\hat{\mu}(\mathcal{S}')\) and \(\hat{\sigma}^2(\mathcal{S}')\), and compute
      \[
          J_j = w_\mu \big(\mu(\mathcal{S}')-\hat{\mu}\big)^2
            + w_\sigma \big(\sigma^2(\mathcal{S}')-\hat{\sigma}^{2\*}\big)^2.
      \]
      \item Select $j^* = \arg\min_j J_j$, set \(\mathcal{S} \leftarrow \mathcal{S}'\), \(T \leftarrow T + |c_{j^*}|\), and
      update \(\hat{\mu}(\mathcal{S}),\hat{\sigma}^2(\mathcal{S})\).
\end{itemize}

\section{Experiment}
\subsection{Experimental Setting}
\paragraph{Hyper-Parameters} We employ the AdamW optimizer with hyper-parameters set to $\beta_1 = 0.9$, $\beta_2 = 0.95$, and weight-decay $= 0.1$. We set the maximum sequence length to 8K during the whole CPT stage. As for the learning rate scheduling, we first use a warming up with peak learning rate of $1 \times 10^{-5}$. During the anneal training procedure, we gradually decay the learning rate following a cosine decay curve. The gradient clipping norm is set to $1.0$. The base model is Qwen3-14B-Base.

\paragraph{Evaluation Datasets}
We evaluate the models on leading benchmarks over both medical and general tasks. Medical tasks include MMLU, DiagnosisArena \citep{zhu2025diagnosisarena}, MedMCQA \citep{pal2022medmcqa}, MedQA-USMLE \citep{jin2021disease}, PubMedQA \citep{jin2019pubmedqa}, MedBullets \citep{chen2025benchmarking}, NEJMQA \citep{katz2024gpt}, SuperGPQA-Med \citep{du2025supergpqa}, GPQA-Med \citep{rein2024gpqa}, and medical subsets of CEVAL \citep{huang2023c}, CMMLU \citep{li2023cmmlu}, MMLU \citep{hendrycks2020measuring}. General tasks include CEVAL, CMMLU, and MMLU.

\paragraph{Baseline} Four baselines are utilized: (1) Base model without CPT, (2) RS-CPT with random data sampling, (3) LPS-CPT with low-PPL sampling, \textit{i.e.}, and (4) HPS-CPT with high-PPL sampling, \textit{i.e.}, PPL scors are lower/higher than optimal one derived from DOS.

\begin{table}[!tb]
\centering
\caption{Effectiveness of Perplexity-Aware Data Scaling Law. All models are evaluated under the same evaluation setting. The highest, the second-best scores are shown in \textbf{bold} and \underline{underlined}, respectively. The base model is Qwen3-14B-Base.}
\label{tab:base-eval}
\small %
\setlength{\tabcolsep}{3pt} %
\begin{tabular}{@{}clccccc@{}}
\toprule
\multirow{2}{*}{\textbf{Task}} & \multirow{2}{*}{\textbf{Benchmark}} & \multirow{2}{*}{\textbf{Base}} & \multicolumn{3}{c}{\textbf{CPT}}\\\cmidrule{4-7}
                               &                                     &                                &  \textbf{RS} & \textbf{LPS} & \textbf{HPS} & \textbf{DOS} \\
\midrule
\multirow{11}{*}{\rotatebox{90}{\textbf{Medical}}}
&DiagnosisArena       & 41.30  & 56.40   & 56.40 &45.70& 61.11 \\
&GPQA-Med             & 57.89  & 57.89   & 57.89 &57.89& 63.16 \\
&PubMedQA             & 76.60  & 76.70   & 76.40 &78.60& 77.40 \\
&Medbullets           & 55.52  & 56.82   & 56.82 &58.44& 57.14 \\
&NEJMQA               & 64.73  & 66.06   & 65.60  &64.79& 66.14  \\
&MedMCQA              & 66.89  & 67.75   & 67.51 &67.75& 68.28  \\
&MedQA-USMLE          & 72.58  & 73.21   & 73.21 &72.66& 73.61  \\
&CEVAL-Med  & 89.86  & 90.14 & 89.63 &91.46& 90.44  \\
&CMMLU-Med  & 86.68  & 86.78 & 86.89 &86.58& 86.38  \\
&MMLU-Med   & 81.19  & 81.62 & 81.85 &81.68& 82.11 \\\cmidrule{2-7}
\rowcolor{gray!20}
&\textbf{Average}    & 69.32 & \underline{71.34}  & 71.22 &70.56& \textbf{72.48} \\
\midrule
\multirow{3}{*}{\rotatebox{90}{\textbf{General}}}&CEVAL    & 85.52    & 85.14  & 85.14 &85.17& 85.44 \\
&CMMLU    & 84.92   & 84.50  & 84.79 &84.53& 84.78 \\
&MMLU     & 82.03   & 82.23  & 82.03 &82.11& 82.25 \\ \cmidrule{2-7}
\rowcolor{gray!20}
&\textbf{Average}  & \textbf{84.16}   & 83.94 & \underline{83.99} &83.94& \textbf{84.16}  \\ 
\bottomrule
\end{tabular}
\end{table}

\subsection{Model Results}
\paragraph{Benchmark Performance} Table \ref{tab:base-eval} evaluates the impact of our perplexity-aware data scaling law on continual pre-training. On medical benchmarks, DOS-Base achieves the best overall performance, reaching an average score of 72.48, compared to 71.34 for RS-CPT and 71.22 for LPS-CPT. This corresponds to a 3.16 improvement over the base model. The gains are consistent across datasets: DOS-Base attains the highest or second-highest score on almost all medical tasks, with particularly notable improvements on DiagnosisArena and GPQA-Med. On general-domain benchmarks, DOS-Base maintains the par-level performance (84.16), matching the base model and slightly exceeding CPT-Base (83.94), indicating that the DOS could well reduce the forgetting of general knowledge. These results demonstrate that guiding CPT with our perplexity-aware scaling law is more effective than naive CPT: it converts additional training data into substantial, domain-specific gains while preserving strong general performance.

\begin{figure}[!t]
    \centering
    \includegraphics[width=1\linewidth]{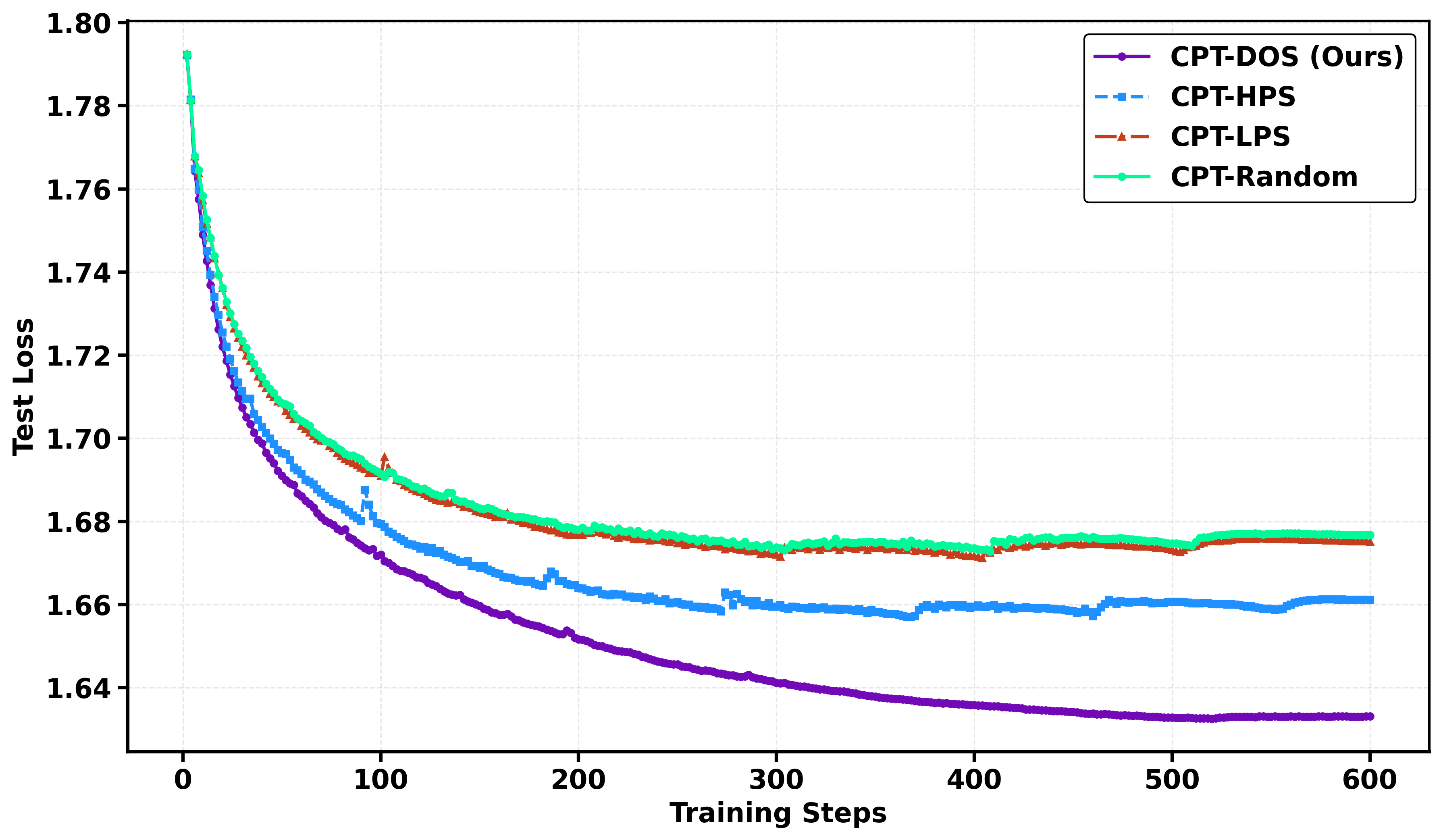}
    \caption{Test loss curves during CPT. CPT with DOS converges more rapidly at a lower loss.}
    \label{fig:loss}
\end{figure}

\paragraph{Loss Curve} As shown in Figure \ref{fig:loss}, across all methods, test losses decrease steadily over the first several hundred training steps, but the four strategies exhibit noticeably different convergence behaviors. CPT‑DOS shows the most rapid and consistent reduction in test loss, achieving the lowest final loss and demonstrating stable improvement throughout training. In contrast, CPT‑HPS converges more slowly and plateaus at a higher loss, exhibiting small fluctuations after roughly 250 steps. CPT‑LPS performs similarly to CPT‑HPS but ultimately settles at a higher loss. CPT‑Random shows the weakest performance, consistently lagging behind all other methods. Overall, the figure highlights that CPT‑DOS not only accelerates convergence but also delivers a substantial improvement in final model performance.

\begin{figure}[!t]
    \centering
    \includegraphics[width=1\linewidth]{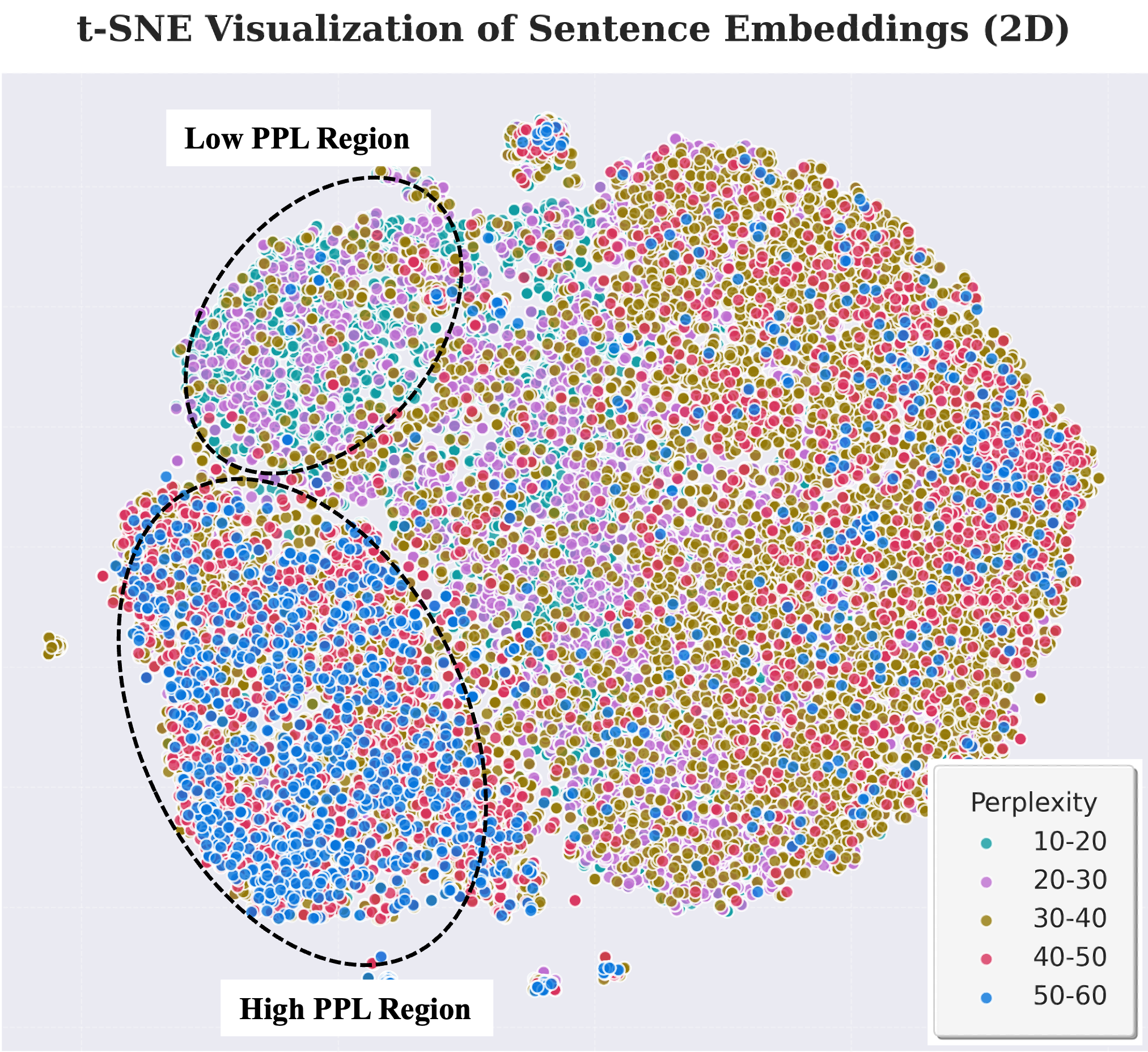}
    \caption{t-SNE visualization for varying perplexity ranges. Two dense subregions are clearly emphasized: a low-PPL region (upper-left) and a high-PPL region (lower-left).}
    \label{fig:tsne}
\end{figure}

\paragraph{Data Distribution} Figure \ref{fig:tsne} presents a t-SNE projection of sentence embeddings, where each point is colored by its perplexity range. Samples with medium PPL values (roughly the middle bins) are more diffusely scattered and less concentrated in distinct clusters. LPS-CPT or HPS-CPT tends to focus on extreme-confidence outputs, while under-representing moderate-difficulty cases, which are often important for balanced generalization. Notably, applying the DOS strategy improves the coverage of the embedding space by explicitly controlling variance during data selection. As a result, DOS produces samples that span both low and high PPL regions, increasing overall data diversity and encouraging exploration of different linguistic patterns reflected in the two extremes. The distribution remains quality-preserving and balanced, suggesting that DOS can avoid bias toward either easy or overly difficult examples while still expanding the effective coverage of the data manifold.

\section{Conclusion}
We rethink the prevailing assumption that model performance scales monotonically with dataset size, indicating that naive data scaling yields diminishing returns during CPT. To address this inefficiency, we introduce a perplexity-aware data scaling law that exploits a model’s initial perplexity landscape over domain corpora as a proxy for knowledge gaps. By modeling the relationship between perplexity statistics and model performance, our method adaptively selects high-utility training subsets while discarding redundant or noisy examples. Empirical results on medical and general-domain benchmarks show that this approach consistently identifies near-optimal data subsets, achieving superior adaptation performance with significantly less data than conventional CPT.

\bibliography{main}

\end{document}